\documentclass[letterpaper, 10 pt, conference]{ieeeconf}  
\pdfoutput=1

\IEEEoverridecommandlockouts                              

\usepackage{graphics} 
\usepackage{epsfig} 
\usepackage{mathptmx} 
\usepackage{times} 
\usepackage{amsmath} 
\usepackage{amssymb}  
\usepackage{graphicx}
\usepackage{booktabs}
\usepackage{multirow}
\usepackage{verbatim}

\usepackage{color} 
\usepackage{xcolor}

\title{\LARGE \bf
	Motion Planning for Heterogeneous Unmanned Systems under Partial Observation from UAV* 
}

\author{Ci Chen$^{1}$, Yuanfang Wan$^{4}$, Baowei Li$^{3}$, Chen Wang$^{2,3}$,  Guangming Xie$^{3,5}$, Huanyu Jiang$^{1}$             
	\thanks{*This work was supported in part by grants from the National Natural Science Foundation of China (NSFC, No.61973007, 61633002). Corresponding author: Chen Wang ({\tt\small wangchen@pku.edu.cn})}
	\thanks{$^{1}$College of Biosystems Engineering and Food Science, Zhejiang University, Hangzhou 310058, China.}%
	\thanks{$^{2}$National Engineering Research Center for Software Engineering, Peking University, Beijing 100871, China.}
	\thanks{$^{3}$The State Key Laboratory of Turbulence and Complex Systems, Intelligent Biomimetic Design Lab, College of Engineering, Peking University, Beijing 100871, China.}%
	\thanks{$^{4}$Department of Ocean Science and Engineering, Southern University of Science and Technology, Shenzhen 518055, China.}
	\thanks{$^{5}$Peng Cheng Laboratory, Shenzhen 518055, China.}
}

\begin{document}
	
	\maketitle
	\thispagestyle{empty}
	\pagestyle{empty}
	
	\begin{abstract}
		For heterogeneous unmanned systems composed of unmanned aerial vehicles (UAVs) and unmanned ground vehicles (UGVs), using UAVs serve as eyes to assist UGVs in motion planning is a promising research direction due to the UAVs' vast view scope. However, due to UAVs flight altitude limitations, it may be impossible to observe the global map, and motion planning in the local map is a POMDP (Partially Observable Markov Decision Process) problem. This paper proposes a motion planning algorithm for heterogeneous unmanned system under partial observation from UAV without reconstruction of global maps, which consists of two parts designed for perception and decision-making, respectively. For the perception part, we propose the Grid Map Generation Network (GMGN), which is used to perceive scenes from UAV's perspective and classify the pathways and obstacles. For the decision-making part, we propose the Motion Command Generation Network (MCGN). Due to the addition of memory mechanism, MCGN has planning and reasoning abilities under partial observation from UAVs. We evaluate our proposed algorithm by comparing with baseline algorithms. The results show that our method effectively plans the motion of heterogeneous unmanned systems and achieves a relatively high success rate. 
	\end{abstract}
	
	\section{INTRODUCTION}
	For heterogeneous unmanned systems composed of UAVs and UGVs, UAVs have the advantages of flight height and broad scope of observation, while UGVs are capable of accurate operation on ground objects. 
	Therefore, the cooperation between UAVs and UGVs can significantly improve the efficiency of task execution.
	Compared with the egocentric view of UGVs, the top-down view of UAVs can more accurately perceive obstacles around UGVs. In the scenarios of unavailable maps, especially during natural disasters or war zones, it becomes convenient to have a UAV that serves as an eye in the sky that explores the environment and generates trajectories guiding the UGVs on the ground in real-time.

\begin{figure}[t]
		\centering
		\includegraphics[width=0.8\linewidth]{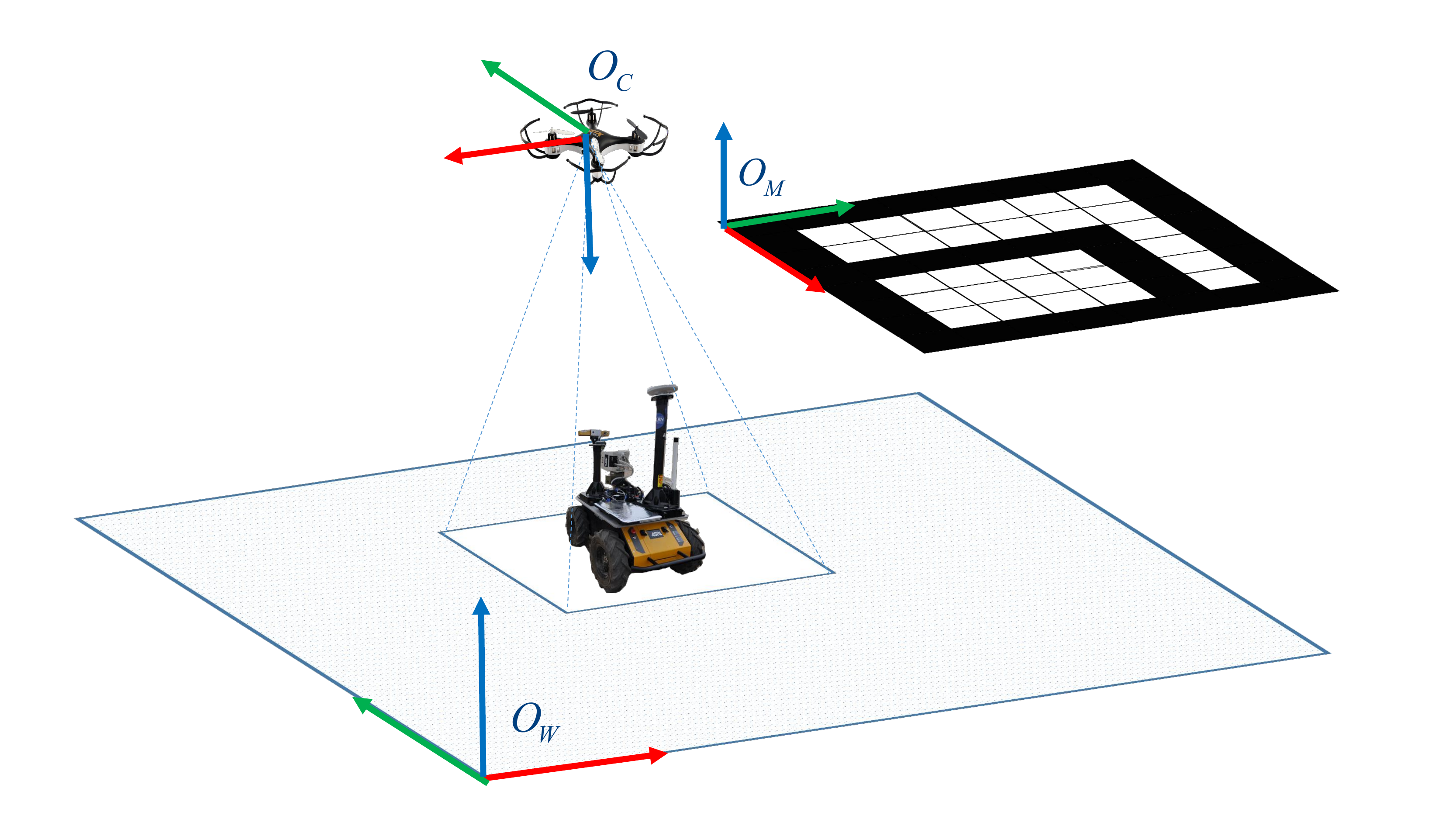}\vspace{-0.2cm}\vspace{-0.2cm}
		\caption{Schematic diagram of partial observation from UAV. $O_{C},O_{M},O_{W}$ represents camera coordinates system, grid map coordinates system, and world coordinates system, respectively. The red (resp. green, blue) arrow represents the $x$ (resp. $y$, $z$) axis.}
		\label{fig_system} \vspace{-0.2cm}\vspace{-0.2cm}\vspace{-0.2cm}
	\end{figure}

	Due to these reasons, some researchers have focused on motion planning for heterogeneous unmanned systems. \cite{01,02,03,04} assume that the UAV's field of vision (FOV) can capture the entire range from starting point to the target point, and planning is carried out as the global environment is known. 
	Unfortunately, in many practical applications, especially when the heterogeneous unmanned systems are exploring in a wide range of unfamiliar environment, only part of the overall environment can be observed by UAV due to its limited perspective (see Fig. \ref{fig_system}).
	To address this, \cite{kashino2019aerial,papachristos2014power,peterson2018online} proposed solutions respectively, all of which need to reconstruct the global map. Unlike the above work, our method doesn't need a reconstruction step, thus saving computation time. 

	Precisely, this algorithm consists of two parts: perception and decision-making. For the former, we propose a generation network GMGN, which can distinguish obstacles from passable parts in the pictures taken by UAV. On this basis, while ensuring navigation accuracy, the segmentation results'size is reduced to guarantee the real-time performance of the subsequent decision-making network.
	For the latter, we propose MCGN. Thanks for the combinations of value iteration network and external memory network, it has the abilities of reasoning and planning simultaneously. Just like humans explore the unfamiliar environment, it can make a relatively accurate prediction of future actions by analyzing the scenes it has seen before and the scenes at present. Its inputs are grid maps generated by the GMGN, and its outputs are discrete motion commands for UAVs and UGVs. 
	This paper's main contribution is to enable the heterogeneous unmanned system to plan the motion under partial observation from UAV without reconstruction of the global map. In our method, the target can be set anywhere, and the heterogeneous unmanned system will learn to find the target. If the target position doesn’t appear within the camera’s FOV, UAV and UGV keep moving synchronously. UAV will remain hovering if the target appears in camera's FOV, and UGV will get to the target alone.

	The rest of this paper is organized as follows.
	Section II reviews related work regarding path planning for heterogeneous unmanned systems and external memory networks.
	Section III presents our proposed algorithm. Then, in Section IV, the experiments in the simulation are delivered to verify the feasibility and accuracy of our proposed algorithm by comparing it with baseline algorithms.
	Finally, Section V concludes this paper and outlines our future work.
	
	\section{RELATED WORK}

	\subsection{Path Planning for Heterogeneous Unmanned Systems}
	
	\cite{01} has demonstrated a fully autonomous collaboration of a UAV and a UGV in a mock-up disaster scenario. The UAV first maps an area of interest, and then it computes the fastest mission for the UGV to reach a spotted victim and provide a first-aid kit. 
	A similar collaborative framework has been proposed in \cite{02}. The UAV collects images of the surroundings firstly. A vision-based algorithm is utilized to recognize roads, pathways and obstacles, and an enhanced version of the A* algorithm is applied to calculate a path around them towards the destination. 
	\cite{03} has developed a hybrid path planning algorithm to optimize the planned path. A genetic algorithm is used for global path planning, and a local rolling optimization is employed to constantly optimize the genetic algorithm results. 
	\cite{04} has proposed a multi-UAVs based stereo vision system to assist global path planning for a UGV even in GPS-denied environments. The proposed method can optimally generate the depth map of ground objects and robustly detect obstacles.
	The above studies assume that the UAV can observe the global environment. When the UAV can only see part of the environment, the following literature explains the planning method of UAV and UGV.
	\cite{kashino2019aerial} proposed a control scheme for collaborative navigation, which relies on an incremental map building strategy proving environment feedback and sampling based trajectory-planning approach (RRT*). 
	Papachristos et al. \cite{papachristos2014power} proposed a method that utilizes target iso-probability curves to plan both UAV and UGV trajectories. For UAVs, the proposed novel search-planning algorithm determines paths that traverse a range of iso-probability curves while covering them with equal effort. Such a search achieves a balance between exploitation and exploration. 
	In \cite{peterson2018online}, the UAV acquires imagery which is assembled into orthomosaic and then classified. These terrain classes are used to estimate relative navigation costs for the ground vehicle, so energy-efficient paths may be generated and then executed. 
	
	\subsection{External Memory Networks}
	
	The neural network is good at pattern recognition and fast-response decision-making, but it isn't good at reasoning. Some research on deep learning uses neural networks with external memory, which can learn from samples like neural networks and store complex data like computers. The combination of the two can realize the fast storage of knowledge and flexible reasoning.
	
	In 2014, Graves et al. \cite{13} have proposed NTM inspired from Turing Machine architecture, which combines neural network with external memory to expand the neural network's capability. It mainly composed of a controller and a memory bank. The neural network is seen as the CPU, and the memory is regarded as the RAM. The controller determines where to read and write information in the memory according to the task. In NTM, a set of vectors is utilized to represent the memory bank. 
	Since the original NTM, there have been some interesting articles exploring similar topics. Neural GPU \cite{14} has solved the problem that NTM cannot perform addition and multiplication operations.
	Zaremba et al. \cite{15} have used reinforcement learning to train NTM to solve simple algorithm problems. 
	Kurach et al. \cite{16} have imitated the practical computer memory which works with a pointer. 
	Some papers have also explored differentiable data structures, such as stacks \cite{17} and queue \cite{18}.
	
	On the bases of NTM, Graves et al. \cite{19} have proposed DNC in 2016, which is the second version of NTM and improves the addressing mechanism of NTM. DNC uses content-based addressing and dynamic memory allocation to handle write operations, content-based addressing and temporary memory linkage to deal with read operations. The DNC can choose how to allocate memory, store information, and easily find data in memory. 
	%
	%
	In this work, due to external memory networks' introduction, our algorithm has reasoning capabilities, which is an important reason for the MCGN to perform navigation under partial observation from UAV.

	\section{APPROACH}
	
	\begin{figure}[t]
		\centering
		\includegraphics[width=0.8\linewidth]{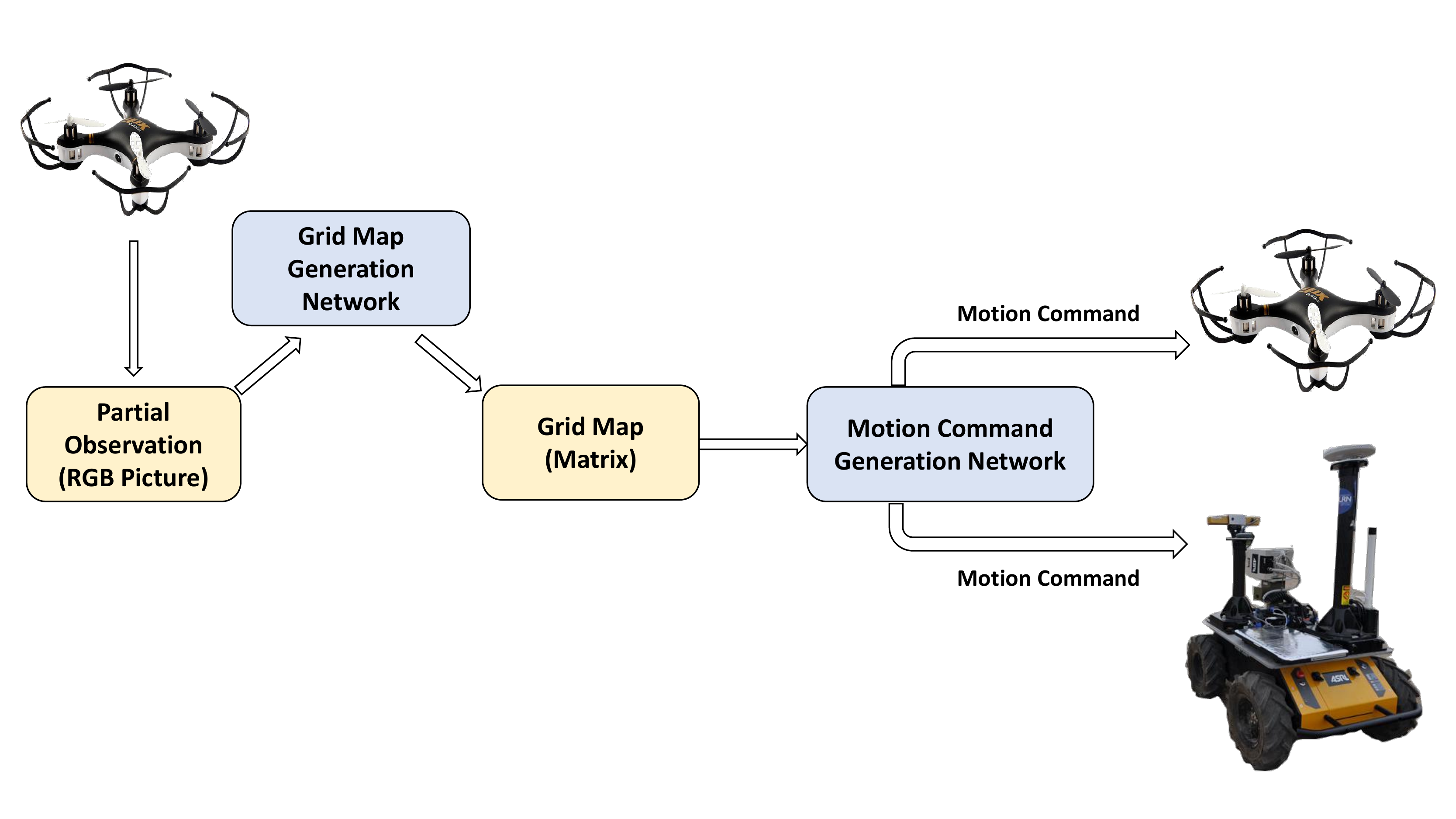}\vspace{-0.2cm}\vspace{-0.2cm}
		\caption{Architecture of our proposed algorithm.}
		\label{fig_architecture}\vspace{-0.2cm}\vspace{-0.2cm}\vspace{-0.2cm}
	\end{figure}
	
	We focus on the heterogeneous unmanned system composed of a UAV and a UGV (see Fig. \ref{fig_system}).
	The UAV flies at a constant height and is equipped with a top-down view camera.
	The UGV has no environmental sensing device. Thus it has to be assisted by the UAV for navigation.
	To provide navigation strategies for such heterogeneous unmanned systems in an unfamiliar environment, we propose an algorithm whose architecture is shown in Fig. \ref{fig_architecture}.
	The algorithm's inputs are parts of the environment observation photographed by the camera equipped on the UAV, and the outputs are discrete control commands of UAV and UGV, including up, down, left and right actions.
	Our algorithm consists of two parts, GMGN and MCGN.
	For GMGN, the inputs are RGB images captured by the camera, and the outputs are grid maps represented by matrices.
	For MCGN, the inputs are grid maps, and the outputs are discrete motion control commands.
	In the following two subsections, these two parts are described in detail, respectively.

	\subsection{Grid Map Generation Network}
	
\begin{figure}[t]
		\centering
		\includegraphics[width=1.0\linewidth]{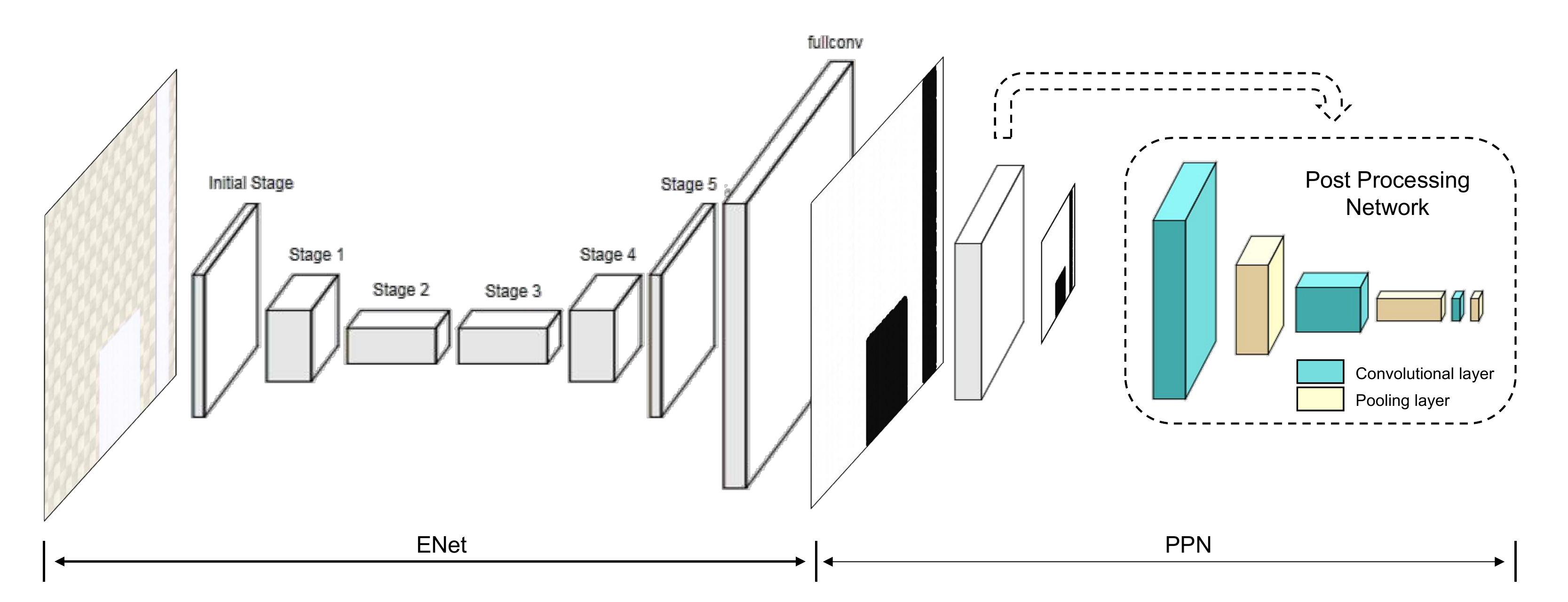}\vspace{-0.2cm}
		\caption{Structure of Grid Map Generation Network}
		\label{fig_GMGN}\vspace{-0.2cm}\vspace{-0.2cm}\vspace{-0.2cm}
	\end{figure}

	GMGN aims to transform RGB images into grid maps.
	Since it’s difficult to directly convert 3-channels images into 1-channel matrices, in GMGN, we first transform RGB images (image\_height, image\_width, channels) into 1-channel semantic segmentation maps (image\_height, image\_width) by using semantic segmentation network, and then reduce the 1-channel semantic segmentation maps to grid maps (matrix\_width, matrix\_height). The structure of GMGN is given in Fig. \ref{fig_GMGN}.
	
	In the semantic segmentation section, we use the fast and lightweight ENet network \cite{12}, which is a real-time segmentation network with considerable accuracy. The network is an encoder-decoder structure, but it is not symmetrical. It uses a larger encoder and a smaller decoder structure.
	Specifically, as shown in Fig. \ref{fig_GMGN},
	the initial stage is comprised of initial blocks, and stages 1-5 are composed of bottleneck modules, where stage 1-3 belong to the encoder and stage 4-5 belong to the decoder.
	
\begin{figure}[t]
		\centering
		\includegraphics[width=1.0\linewidth]{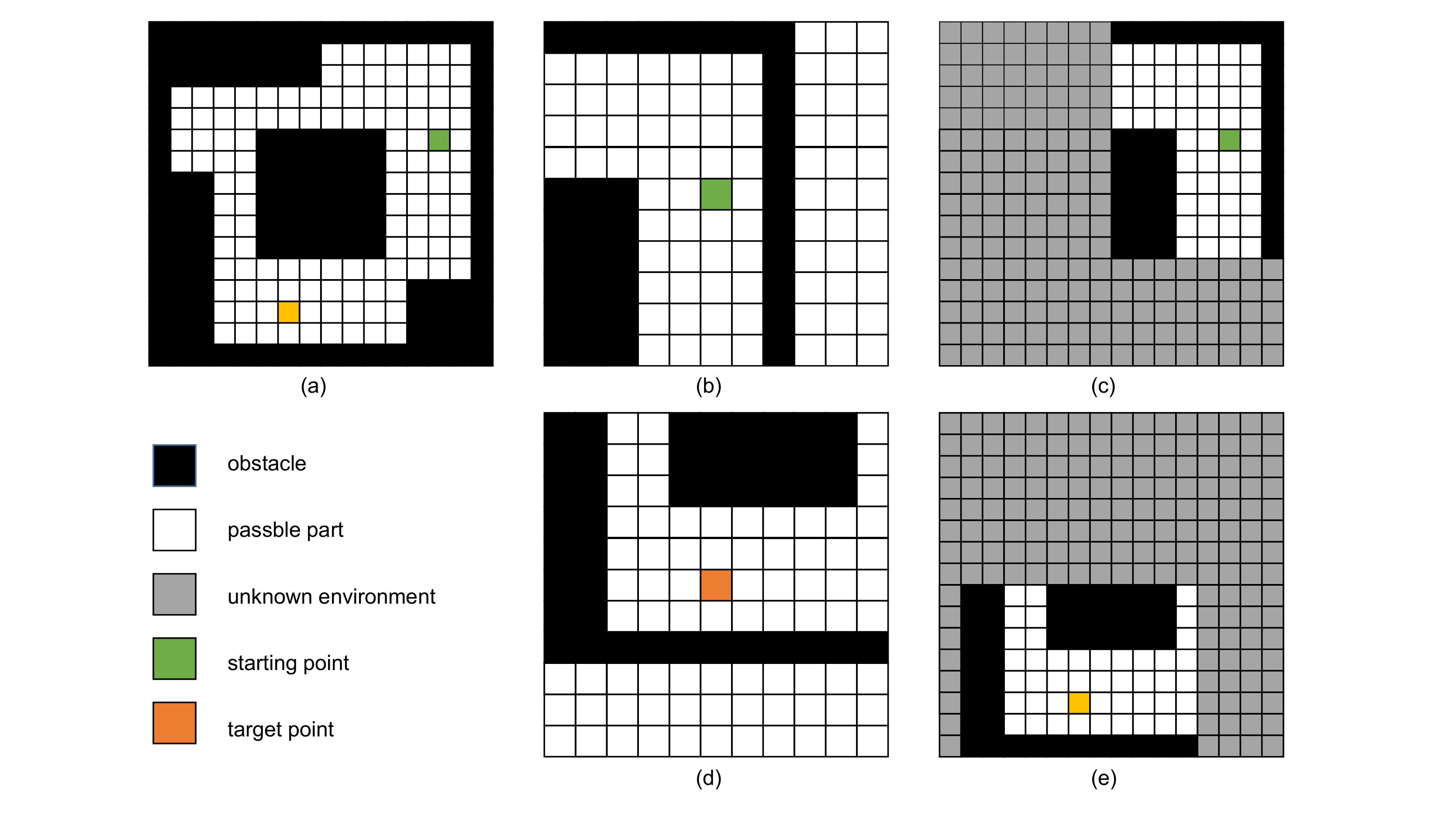}\vspace{-0.2cm}
		\caption{Explanation of the grid map. (a) The global observation $O_{a}$ ranges from the starting point to the target point. (b) The partial observation matrix $O_{p}$ when UAV at the starting point. (c) The masked global observation matrix $O_{g}$ when UAV at the starting point. (d)The partial observation matrix $O_{p}$ when the target point appears in UAV’s FOV. (e) The masked global observation matrix $O_{g}$ when the target point appears in the UAV’s FOV.}\vspace{-0.2cm}
		\label{fig_gridmap}
	\end{figure}

	We then design the Post Processing Network (PPN) to transform the semantic segmentation maps into grid maps.
	As shown in Fig. \ref{fig_gridmap}, grid map is a two-dimensional matrix, also known as the partial observation matrix $O_{p}$. The size of $O_{p}$ is $m\times n$, which represents the scene observed by UAV’s camera. 
	The starting point is represented as an orange grid, and the target point is a green grid.
	Based on these, a masked global observation matrix $O_{g}$ with the size of $M\times N (M>m,N>n)$ can be generated, which reflects global information. In this matrix $O_{g}$, the unknown environment (outside the camera’s FOV) is set to $-1$, while $0$, $1$, and $2$ represent the passable parts, the obstacles, and the target in the known environment, respectively. Note that a target in the known (resp. unknown) environment is set to $2$ (resp. $-1$).

	\subsection{Motion Command Generation Network}
	
	In this subsection, we will first introduce the VIN module and DNC module principles in brief, which are important components in MCGN, and then analyze the network structure of MCGN in detail.

	\subsubsection{Value Iteration Network}
	
	Path planning is a sequential decision problem, that can be regarded as a Markov Decision Process (MDP). MDP is a 5-tuple model 
	$\left\langle S,A,P,R,\gamma \right\rangle$,
	where $S$ is the masked global observation matrix and stands for a set of states, $A$ is discrete motion command and stands for actions,  $P$ presents the conditional transition probabilities between states, $R$ means the reward function and $\gamma$ is the discount factor.
	Strategy $\pi(a|s)$ determines the probability of taking actions $a$ under states $s$. 
	The goal is to maximize long-term returns. The value iteration method is a traditional method to solve the MDP problem. The state value function $V(s)$ can be obtained by iteratively calculating the state-action value function $Q(s,a)$ of each state (Eq. (1)). By iterating all possible actions in each state many times, we can get the optimal strategy (Eq. (2)).
	\begin{eqnarray}
	Q_{n}(s,a)=R(s,a)+\sum_{s^{'}}\gamma P(s^{'}|s,a)V_{n}(s^{'}), \\\nonumber
	\text{where}\ V_{n+1}(s)=\max_{a}Q_{n}(s,a), \;\;\; \forall s \\
	\pi^*(a|s)=\arg \max_{a}Q^*(s,a)
	\end{eqnarray}
	
\begin{figure}[t]
		\centering
		\includegraphics[width=0.8\linewidth]{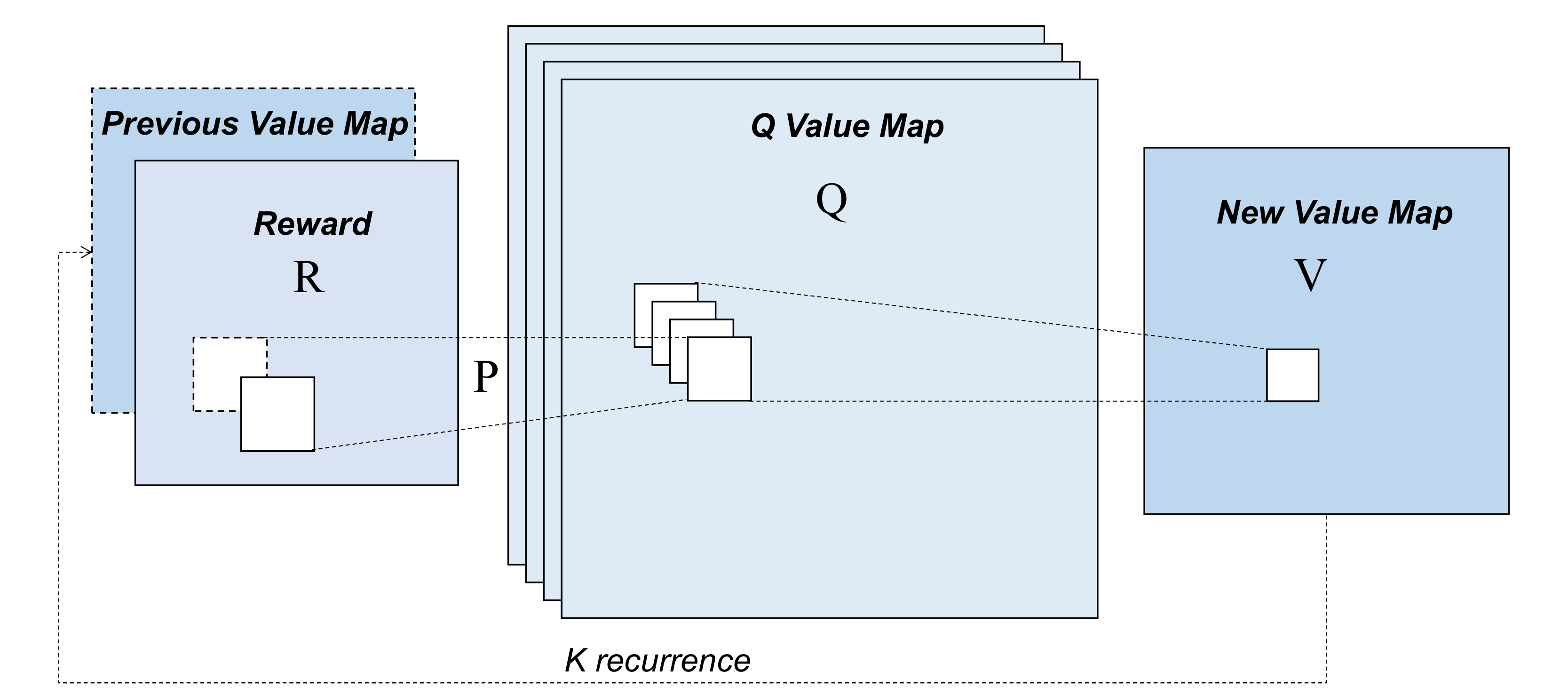}\vspace{-0.2cm}
		\caption{Value Iteration Network Module}
		\label{fig_vin}\vspace{-0.2cm}
	\end{figure}

	We adopt a network called value iteration network (VIN) module \cite{21}, as shown in Fig. \ref{fig_vin}. This structure has the same mathematical expression as the classical programming algorithm value iteration mentioned above. In this network, value iteration is expressed as a convolutional neural network (CNN), which is differentiable. So the whole network can be trained using standard backpropagation. This makes VIN module simple to train using imitation learning (IL) or reinforcement learning (RL) algorithms, and straightforward to integrate with NNs for perception and control.
	
	\subsubsection{Differential Neural Computer}
	DNC is a exclusive recurrent neural network (RNN) with external memory. It uses vectors to store memory. Each row of the memory matrix $M\in R^{N \times M}$ corresponds to a unique memory. The controller utilizes interface vector to control one write head and multiple read heads to interact with external memory.
	External memory matrix update as
	\begin{equation}
	M_{t}=M_{t-1}(E-w_{t}^{W}e_{t}^{T})+w_{t}^{W}v_{t}^{T}
	\end{equation}
	where $M_{t}$ is the memory matrix at time step $t$, $E\in R^{N \times M}$ is the identity matrix, $e_{t}^{T}\in R^{W}$ is the erase vector, and $v_{t}^{T} \in R^{W}$ is the write vector. $w_{t}^{W} \in R^{N}$ is the write weight obtained through two addressing mechanisms, content-based addressing and dynamic memory allocation. 
	As shown in Eq. (3), the external memory matrix is erased first and then written, and then the memory matrix is updated.
	
	The read operation is defined as a weighted average over the content of the memory matrix. It produces a set of vectors defined as read vectors as follows
	\begin{equation}
	re_{t}^{i}=M_{t}^{T}w_{t}^{read,i}
	\end{equation}
	where $re_{t}^{i}$ is the read vector which will be appended to the next time step of controller to provide access to memory, $w_{t}^{read,i}$ is the read weight obtained through two addressing mechanisms, content-based addressing and temporal memory linkage.
	
	\subsubsection{Motion Command Generation Network}
	
	Through GMGN, the RGB images taken by the UAV are converted to grid maps. The grid map is known as the partial observation matrix $O_{p}$. The masked global observation matrix $O_{g}$ can be obtained through $O_{p}$ and the current UAV's position. Here, we will introduce how to plan the movement of a heterogeneous unmanned system through $O_{g}$ and $O_{p}$.
	
	When planning starts, the actions of UAV and UGV are synchronized, and the UGV remains in the center of the UAV's FOV. Assuming that the target point appears in the UAV's FOV at time step $\tau$. At time step 0 $\sim$ $\tau$, inspired by the Memory Augmented Control Networks (MACN) \cite{22}, we use VIN to learn motion planning and use DNC to track and record import signs of environments. The combination of VIN and DNC can determine the optimal strategy in the global environment by collecting strategies calculated in the partial observation space.
	After time step $\tau$, UAV keeps hovering, and UGV will get to the target point alone. At this time step, for heterogeneous unmanned system, camera's observation can accurately model the current environment. Therefore, it is possible to plan the motion by VIN without resorting to DNC directly.
	
	\begin{figure*}[thpb]
		\centering
		\includegraphics[width=0.8\linewidth]{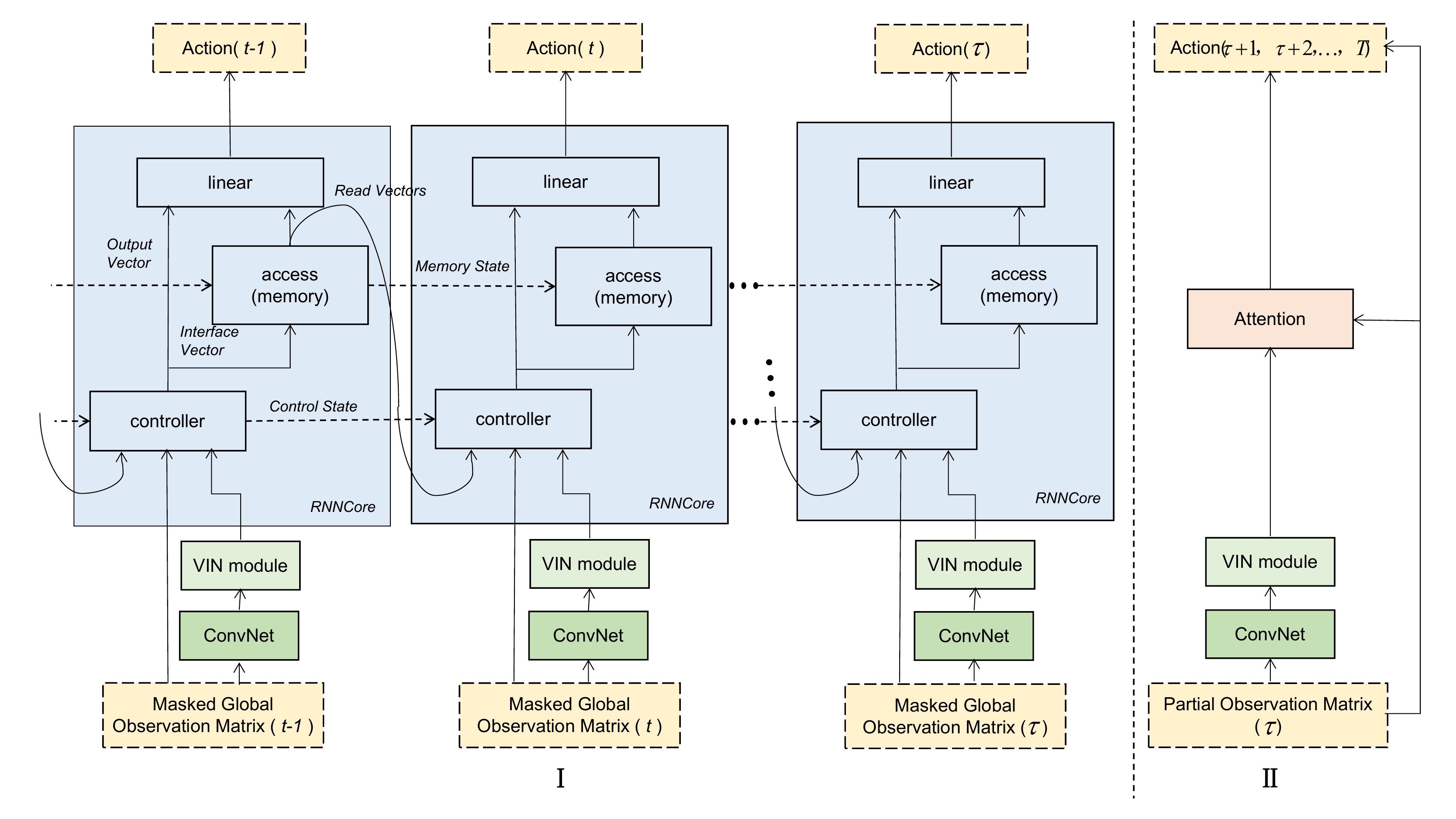}\vspace{-0.2cm}
		\caption{Structure of Motion Command Generation Network}
		\label{fig_MCGN}\vspace{-0.2cm}\vspace{-0.2cm}\vspace{-0.2cm}
	\end{figure*}

	As shown in Fig. \ref{fig_MCGN}, the inputs of the MCGN are the masked global observation matrices $O_{g}$ or the partial observation matrices $O_{p}$, and the outputs are the discrete motion commands of UAV and UGV. The whole network includes ConvNet module, VIN module, Attention module and DNC module which is implemented as a collection of RNNCore modules. 
	The structure of ConvNet module is shown in Tab. \ref{Tab_convnet}. 
	At time step 0, the initial reward map is designed manually, where the value of the target point is set to 10 which represents the reward, and the other values are set to 0.
	At time step 0 $\sim$ $\tau$, the masked global observation matrix and the initial reward map are utilized to form a multi-dimensional matrix with the size of $[M \times N \times 2]$ as the input of ConvNet module. ConvNet module outputs reward map $R$ and value map $V(s)$, both of them are input into the VIN module, after $k$ times of iterations, the final value map is obtained as the input of controller in DNC. 
	
	\begin{table}[htbp] 
		\centering
		\caption{\label{Tab_convnet}Structure of ConvNet} \vspace{-0.2cm}
		\setlength{\tabcolsep}{0.5mm}{
			\begin{tabular}{cccccc} 
				\toprule 
				Type & Filter\_size & In\_channels & Out\_channels & Remarks \\ 
				\midrule 
				Conv\_1 & (3,3) & 2 & 150  & $\backslash$   \\ 
				Pooling & (1,1) & 150 & 1  & Reward map $R$\\ 
				Conv\_2 & (3,3) & 1 & 4  & Action value function $Q(s,a)$\\ 
				max & $\backslash$ & $\backslash$ & $\backslash$ & Value map $V(s)$ \\ 
				\bottomrule 
		\end{tabular}}\vspace{-0.2cm}
	\end{table}
	
	At each time step $t$, as shown in Fig. \ref{fig_MCGN} (\uppercase\expandafter{\romannumeral1}), the input of the controller is the masked global observation matrix $O_\mathrm{g}$, the final value map $V(s)$ generated by the VIN module, as well as the read vectors $r_{t-1}$ and control state generated at time step $t-1$. The controller emits interface vector $\xi_{t}$ and output vector $v_{t}$ after calculation. First, the interface vector $\xi_{t}$ controls the access module update write head, then the write head updates the memory matrix through erasing and writing operations, after that, the read head is updated, finally the read head obtains the read vector $r_{t}$ at time step $t$ based on the new memory matrix. The read vector $r_{t}$ and output vector $v_{t}$ are mapped to the output after linear transformation, that is, discrete action values. $action \in {\left\lbrace 0,1,2,3\right\rbrace }$, corresponding to the four actions of up, down, left and right respectively. $r_{t}$ is passed as new short-term memory to the next time step $t+1$. The DNC's final output is obtained by linear combination of read vector and output vector, similar to the ``Search of Associative Memory'' model of long-term and short-term memory in the human brain.
	
	When the target appears in the FOV of the camera, for the heterogeneous unmanned system, the perspective of the camera can accurately model the current environment. Therefore, the movement can be directly planned by VIN without resorting to DNC. As shown in Fig. \ref{fig_MCGN} (\uppercase\expandafter{\romannumeral2}), the input of the algorithm is the partial observation matrix $O_{p}$ instead of the masked global observation matrix $O_{g}$. The partial observation matrix is input into ConvNet module to generate a reward map and a value map, and both of them are input into a VIN module. After $k$ times of iterations, the final value map and the final reward map can be obtained. These two maps can be added into the attention module, and finally a series of actions to the target point is obtained. Unlike the above, at this time, the UAV keeps hovering, and the commands are only used to control the movement of UGV.
	
	\section{Experiments and Results}
	
	\subsection{Implementation details}
	
	\subsubsection{Experimental Setup}
	To evaluate the effectiveness of the proposed approach, we conducted experiments on the platform HeROS \cite{23}. HeROS is a simulation platform for heterogeneous unmanned system which is developed based on {\tt\small V-rep} software. In this platform, the underlying control programs for UGV and UAV are defined in embedded scripts. 
	UAV and UGV each correspond to a virtual control point. The commands issued by our proposed algorithm first control the virtual point to move one step length through the Remote API, and then the UAV (or UGV) tracks the virtual point through the control program written in the embedded script.
	
	Our experiments were conducted on the following specifications: i5-9400F CPU, 16GB RAM, and NVIDIA GeForce RTX 2060 GPU. The implementation of neural networks has been carried out with Python and Tensorflow library.

	\subsubsection{Grid Map Genetation Network}
	For each RGB image captured by the UAV, its corresponding semantic segmentation map and grid map need to be labeled. When labeling the former, we label obstacles as 1 and passable parts as 0 with the help of the {\tt\small LabelMe} software. 
	When labeling the latter, obstacles are also labeled as 1 and passable parts are labeled as 0 to obtain the grid map and saved in the {\tt\small .npy} format. During training, ENet and PPN are trained separately. In the first place, we train the ENet using RGB images and semantic segmentation maps. 
	In the second place, we use the semantic segmentation maps obtained in the previous step and the manually labeled grid maps to train the PPN. After the two networks converge separately, we combine both networks and train together to fine-tune the parameters.
	
	\begin{table}[t] 
		\centering
		\caption{\label{Tab_paraMCGN}Parameters of Motion Command Generation Network} \vspace{-0.2cm}
		\setlength{\tabcolsep}{2mm}{
			\begin{tabular}{cc} 
				\toprule 
				Parameters & Value \\ 
				\midrule 
				Learning rate & 10e-5   \\ 
				Number of epochs for training & 120 \\
				Number of episodes per epochs & 200 \\
				Batch size & 64 \\
				Number of iteration for planning (VIN) & 30 \\
				Channels in Q layers (actions)  & 4 \\
				Channels in initial hidden layers & 150 \\
				Size of LSTM hidden layers & 256 \\
				Number of memory slots & 32 \\
				Width of each memory slot & 8 \\
				Number of memory read heads & 4 \\
				Number of memory write heads & 1 \\
				\bottomrule 
		\end{tabular}}\vspace{-0.2cm}\vspace{-0.2cm}\vspace{-0.2cm}
	\end{table}

	\subsubsection{Motion Command Generation Network}
	To accelerate the training process, we adopt the imitation learning method, learning from decision data provided by human experts. For each $O_{a}$ (as shown in Fig. 4(a)), an optimal actions sequence from the starting point to the target point is generated by using the A* algorithm firstly. Combined with the current position of the UAV, we retain the value of the observation range and set the value outside the observation range to $-1$, and then we can get the masked global observation matrix $O_{g}$ at this time step (as shown in Fig. 4(c), (e)). 
	Each $O_{g}$ corresponds to an action, which is the optimal action under current observation. 
	
	
	Since the network structure changes before and after time step $\tau$, we train MCGN-\uppercase\expandafter{\romannumeral1} and  MCGN-\uppercase\expandafter{\romannumeral2} separately (as shown in Fig. 6). When the two models have converged, we utilize the two trained models in parallel. When the target point doesn't appear in the UAV's FOV, we use model \uppercase\expandafter{\romannumeral1}. When the target point appears, we use model \uppercase\expandafter{\romannumeral2}. The network parameter design is shown in Tab. \ref{Tab_paraMCGN}.

	\subsubsection{Coordinate Translate}
	Knowing the heterogeneous system's origin position and motion command, we can get the coordinates of the UAV and the UGV under the grid map at each time step. When the algorithm is applied in a simulation environment or a real scene, the coordinates under the grid map need to be transformed into the coordinates in the corresponding scene. This involves three coordinates systems,  the grid map coordinates system, the camera coordinates system, and the world coordinates system (see Fig. \ref{fig_system}).
	
	For a point $P$,  the relationship between its coordinates in the world coordinates system and the coordinates in the camera coordinates system is as follows	
	\begin{eqnarray}
	& {}^CP = {}^C{T_W}{}^WP\\
	& {}^C{T_W} = \left[ {\begin{array}{*{20}{c}}
	{\cos \pi }&0&{\sin \pi }&{x_W^C}\\
	0&1&0&{y_W^C}\\
	{ - \sin \pi }&0&{\cos \pi }&{z_W^C}\\
	0&0&0&1
	\end{array}} \right]
	\end{eqnarray}
	where ${}^C{T_W}$ is the pose transformation matrix between the world coordinates system and the camera coordinates system, ${}^CP$ (resp. ${}^WP$) is the position coordinate of point $P$ in the camera (resp. world) coordinates system, $x_W^C,y_W^C,z_W^C$ represent the position coordinates of the camera coordinates system's origin in the world coordinates system.
	
	Similarly, for point $P$, the relationship between its coordinates in the camera coordinates system and the coordinates in the grid map coordinates system is as follows
	\begin{eqnarray}
	& {}^MP = {}^M{T_C}{}^CP\\
	& {}^M{T_C} = \left[ {\begin{array}{*{20}{c}}
	{\cos \frac{\pi }{2}}&{ - \sin \frac{\pi }{2}}&0&{x_C^M}\\
	{\cos \pi \sin \frac{\pi }{2}}&{\cos \pi \cos \frac{\pi }{2}}&{ - \sin \pi }&{y_C^M}\\
	{\sin \pi \sin \frac{\pi }{2}}&{\sin \pi \cos \frac{\pi }{2}}&{\cos \pi }&{z_C^M}\\
	0&0&0&1
	\end{array}} \right]
	\end{eqnarray}
	where ${}^M{T_C}$ is the pose transformation matrix between the camera coordinates system and the grid map coordinates system, ${}^MP$ is the position coordinate of the point $P$ in the grid map coordinates system, and $x_C^M,y_C^M,z_C^M$ represent the position coordinates of the grid map coordinate system's origin in the camera coordinate system.
	
	From Eq. (5) and (7), the transformation relationship between the coordinates of point $P$ in the world coordinates system and grid map coordinates system can be obtained as
	\begin{eqnarray}\nonumber  
	&{}^WP = {}^M{T_W}^{ - 1} \times {}^MP,\\
	& \text{where}\ {}^M{T_W} = {}^M{T_C} \times {}^C{T_W}.
	\end{eqnarray}

	\subsection{Traditional Baselines}
	The traditional A* algorithm must know the global observation matrix $O_{a}$, only know the partial observation matrix $O_{p}$ is unable to do planning. In order to make the A* algorithm can be applied in the partial observation environment proposed in this paper, we proposed two improved A* algorithms. The principle of Traditional Baselines 1 and Traditional Baselines 2 are shown in Tab. \ref{tab_baseline_1} and Tab. \ref{tab_baseline_2} respectively, where \emph{SearchMinDis2Target()} means to search the point nearest to the target point in $O_{p}$, \emph{AStar()} means to search for a path from starting point to target point using traditional A* algorithm.
	
	\begin{table}[t] 
		\centering
		\caption{\label{tab_baseline_1}Traditional Baselines 1}
		\setlength{\tabcolsep}{2mm}{
			\begin{tabular}{l} 
				\toprule 
				\textbf{Input:} $O_{p}$, Target, Start \\
				\textbf{Output:} ActionSeq  \\
				\midrule 
				MidPoint = \emph{SearchMinDis2Target($O_{p}$, Start, Target)}  \\ 
				ActionSeq = \emph{AStar(map:$O_{p}$, start:Start, goal:MidPoint)} \\
				Execute ActionSeq for UAV and UGV , get the new $O_{p}$ \\
				CurrentPos = MidPoint \\
				\textbf{while} Target not in the $O_{p}$: \\
				\quad MidPoint = \emph{searchMinDis2Target(new $O_{p}$, CurrentPos, Target)} \\
				\quad ActionSeq = \emph{AStar(map:new $O_{p}$, start:CurrentPos, goal: MidPoint)} \\
				\quad Execute ActionSeq for UAV and UGV, get the new $O_{p}$ \\
				\quad CurrentPos = MidPoint \\
				ActionSeq = \emph{Astar(map:new $O_{p}$, start:CurrentPos, goal:Target)}\\
				Execute ActionSeq for UAV and UGV \\
				\bottomrule 
		\end{tabular}}
	\end{table}
	
	\begin{figure}[t]
		\centering
		\includegraphics[width=1 \linewidth]{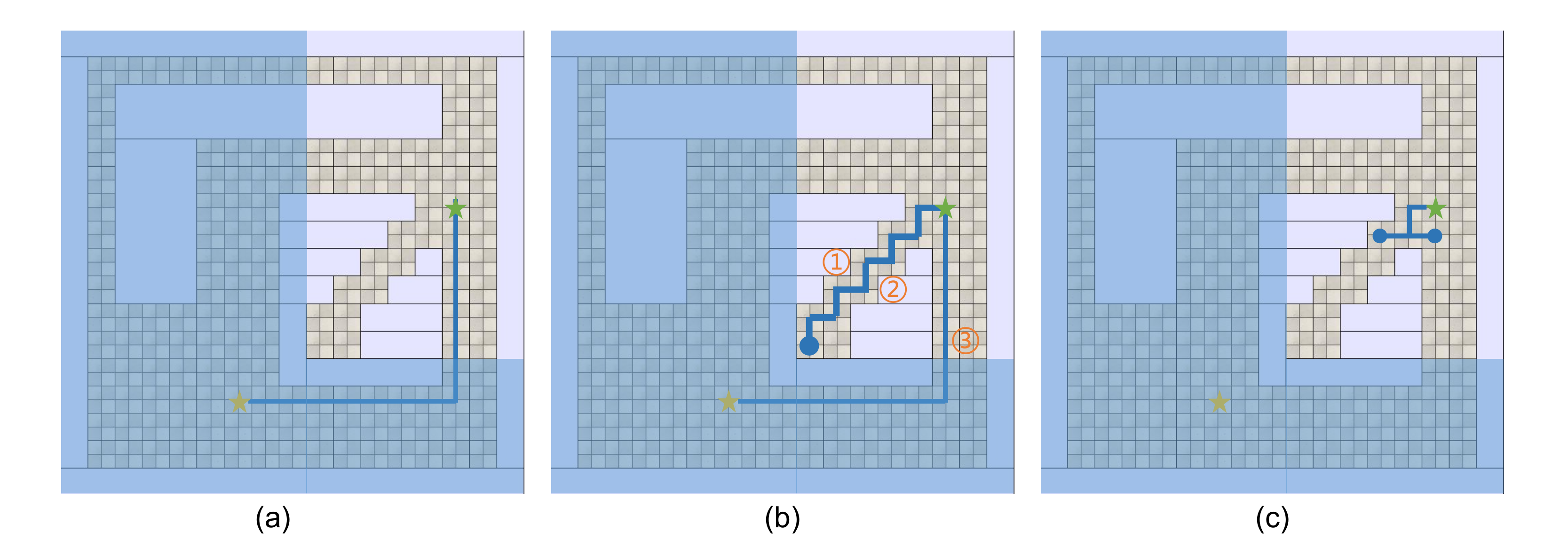}\vspace{-0.2cm}
		\caption{Comparison with baselines}
		\label{fig_baseline}\vspace{-0.2cm}\vspace{-0.2cm}
	\end{figure}

	\begin{table}[t] 
		\centering
		\caption{\label{tab_baseline_2}Traditional Baselines 2} \vspace{-0.2cm}
		\setlength{\tabcolsep}{2mm}{
			\begin{tabular}{l} 
				\toprule 
				\textbf{Input:} $O_{p}$, Target, Start \\
				\textbf{Output:} Action \\
				\midrule 
				MidPoint = \emph{SearchMinDis2Target($O_{p}$, Start, Target)}  \\ 
				ActionSeq = \emph{AStar(map:$O_{p}$, start:Start, goal:MidPoint)} \\
				Action = ActionSeq[0] \\
				Execute Action for UAV and UGV , get the new $O_{p}$ \\
				CurrentPos = MidPoint \\
				\textbf{while} Target not in the new $O_{p}$: \\
				\quad MidPoint = \emph{searchMinDis2Target(new $O_{p}$, CurrentPos, Target)} \\
				\quad ActionSeq = \emph{AStar(map:new $O_{p}$, start:CurrentPos, goal: MidPoint)} \\
				\quad Action = ActionSeq[0] \\
				\quad Execute Action for UAV and UGV, get the new $O_{p}$ \\
				\bottomrule 
		\end{tabular}}\vspace{-0.2cm}\vspace{-0.2cm}
	\end{table}

	\subsection{Results and Analysis}
	
	\subsubsection{Comparison with traditional baselines}
	In order to verify the advantages of MCGN, we selected the following maps to ascertain the effectiveness of MCGN and the baseline algorithms.

	The experimental results are shown in Fig. \ref{fig_baseline}, where the light blue area represents the part which is outside the FOV of UAV at the initial moment, the green star represents the starting point, the yellow star represents the target point and the solid blue line represents the planned path. 
	Fig. \ref{fig_baseline}(a) shows the path planned by MCGN, it can be seen that the algorithm successfully avoids the ``trap'' area and finds the globally optimal path. 
	Fig. \ref{fig_baseline}(b) shows the path planned by Traditional baseline 1. In the initial view of UAV, the blue dot is the position closest to the target point. Therefore, under the current local map, the planned path is from the starting point to the blue dot. When the UAV and UGV reach the blue dot position, the target point appears in the UAV’s FOV, and there is no road at the lower left. In order to reach the target point, the UGV needs to turn back to the starting point, and then go to the target point. 
	Fig. (c) shows the results of the Traditional baseline 2. This method fails under the current map, which plans a circular decision of left and right.
	The above experiments prove that the traditional method doesn't work well under the condition that the global map is unknown.

	\subsubsection{comparison with MACN}
	
	Memory Augmented Control Networks (MACN) \cite{22} is designed for planning problems in partially observable environments, and our method is an improvement on the basis of MACN. In order to evaluate the advantages of our method compared to MACN, we take MACN as the baseline and conduct an evaluation between them. We designed three tasks with different scales. Among them, Task1 and Task2 have the same global scale (the size of the masked global observation matrix), but the observation scope (the size of the partial observation matrix) is different. The observation scope of Task2 and Task3 is uniform, but the global scales are different. In our experimental design, the parameters are summarized in Tab. \ref{Tab_parameters}.
	
	\begin{table}[htbp] 
		\centering
		\caption{\label{Tab_parameters}Simulator Parameters} 
		\setlength{\tabcolsep}{2mm}{
			\begin{tabular}{cc} 
				\toprule 
				Parameters & Value \\ 
				\midrule 
				Global scale (M, N) of Task1 & $17\times17$ \\
				Observation range (m, n) of Task1 & $11\times11$ \\
				Global scale (M, N) of Task2 & $17\times17$ \\
				Observation range (m, n) of Task2 & $9\times9$ \\
				Global scale (M, N) of Task3 & $15\times15$ \\
				Observation range (m, n) of Task3 & $9\times9$ \\
				\bottomrule 
		\end{tabular}}
	\end{table}
	
	In Task1, Task2 and Task3, we added different numbers and different sizes of obstacles in each scenario, where the position of obstacles, starting point and target point were randomly set. Besides, we detected whether there is a feasible path between the starting point and the target point, and eliminated the scenario where there is no possible path. For the three tasks, we conducted 400 experiments in each task, of which 100 experiments were performed in the scenarios with 2, 3, 4 and 5 obstacles, respectively. The results are summarized in Tab. \ref{Tab_results}.
	
	In addition, we calculated the average time taken by our proposed method. The GMGN took about 0.827s to process a picture of 256$pixels$ $\times$ 256$pixels$, and the MCGN issued a motion control command in about 0.009s. Therefore, our method meets the real-time requirements in practical applications.
	
	\begin{table}[t]
		\centering
		\caption{\label{Tab_results}Results of accuracy comparison experiments}\vspace{-0.2cm}
		\begin{tabular}{cp{4em}p{3em}p{3em}p{3em}p{3em}}
			\toprule
			\multicolumn{1}{p{4em}}{Task} & Method & Obs\_2 & Obs\_3 & Obs\_4 & Obs\_5 \\
			\midrule
			\multicolumn{1}{c}{\multirow{2}[2]{*}{Task1}} & MACN  & 84/100 & 70/100 & 70/100 & 72/100 \\
			& Ours & 93/100 & 85/100 & 84/100 & 85/100 \\
			\midrule
			\multicolumn{1}{c}{\multirow{2}[2]{*}{Task2}} & MACN  & 79/100    & 65/100    & 64/100    & 60/100 \\
			& Ours & 88/100    & 83/100    & 79/100    & 77/100 \\
			\midrule
			\multicolumn{1}{c}{\multirow{2}[2]{*}{Task3}} & MACN  & 82/100    & 68/100    & 65/100    & 64/100 \\
			& Ours & 90/100    & 82/100    & 82/100    & 79/100 \\
			\bottomrule
		\end{tabular}%
		\label{tab:addlabel}\vspace{-0.2cm}\vspace{-0.2cm}\vspace{-0.2cm}%
	\end{table}%
	
	From Tab. \ref{Tab_results} we can see that for the three tasks with different global scales and observation ranges, our method improves the accuracy rate by about 14\% compared with MACN. Define the observation ratio as the observation range divided by the global scale, we can get the observation ratio of Task1 is 0.65, the observation ratio of Task2 is 0.53, and the observation ratio of Task3 is 0.6. Through the comparison of the data in Tab. \ref{Tab_results}, we can see that the larger the observation ratio, the higher the accuracy of the task. It's easy to explain, when exploring in the unfamiliar environment, the larger the scope of our surrounding environments, the easier it is to plan the path.
	
	Then we compared the scenario with different obstacles in the same task. We can find that when the number of obstacles is 2, the accuracy of both methods is higher. When the number of obstacles exceeds 3, as the number of obstacles increases, the accuracy of both methods is not significantly reduced. Through inspection of the simulation environment, we found that the reason is that when the number of obstacles exceeds 3, the number of obstacles that can be seen by the camera remains basically the same due to the limited field of view, which leads to the above results.
	
	\begin{figure}[thpb]
	\setlength{\abovecaptionskip}{0cm}   
	\setlength{\belowcaptionskip}{0.2cm}   
	\centering
	\includegraphics[width=0.9\linewidth]{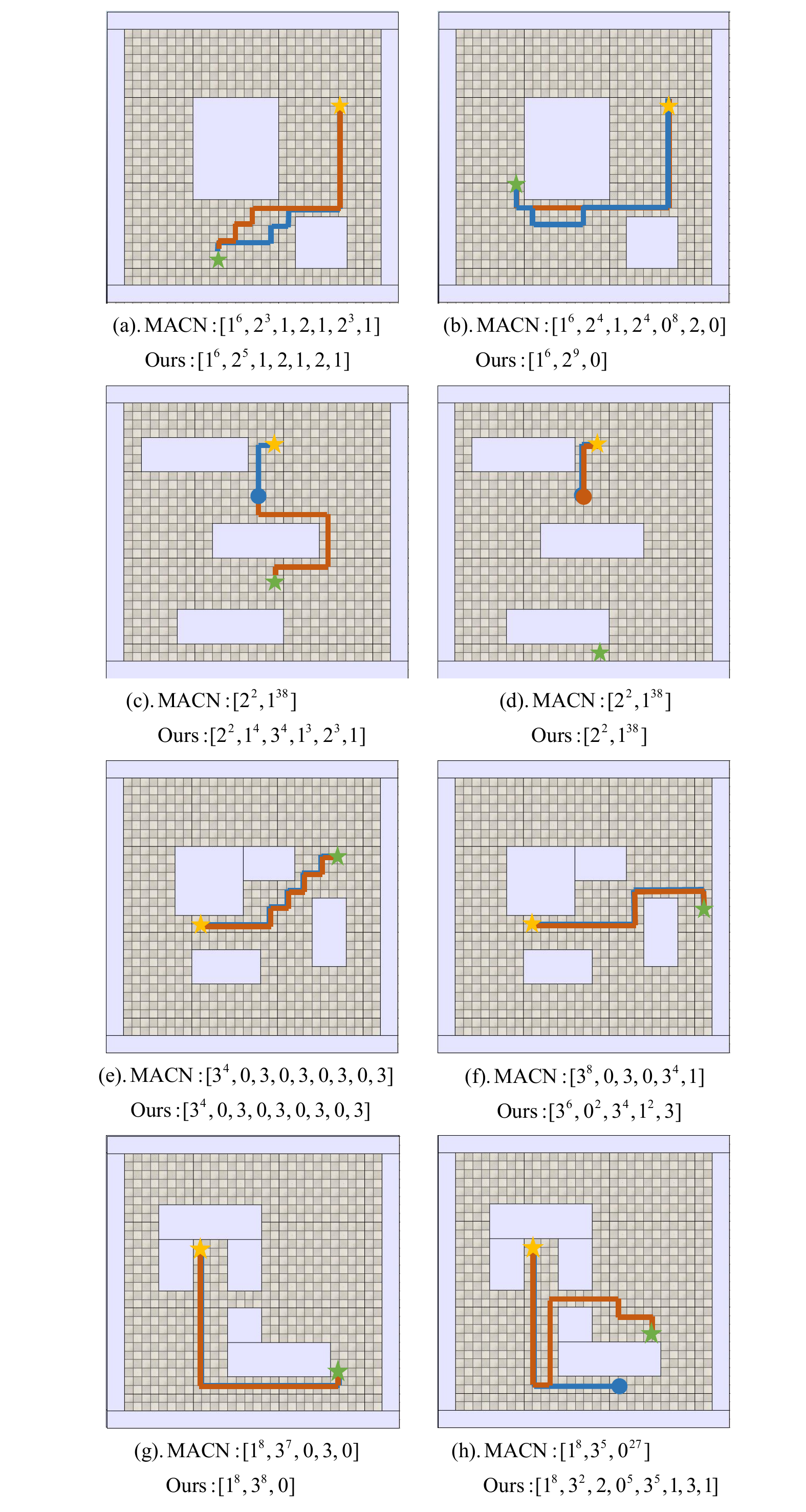}
	\caption{Comparison of the two methods under Task1, where (a), (b) are scenarios with two obstacles, they have the same starting point and different target point. And (c), (d), (e), (f), (g), (h) can be deduced from (a), (b). The red line represents the path planned by our method, the blue line represents the path planned by MACN, the yellow star represents the starting point, the green star represents the target point. Below each picture are the commands output by the two algorithms, the superscript indicates the number of times.}
	\label{fig_macn_ours} \vspace{-0.2cm} \vspace{-0.2cm}
	\end{figure}
	
	To further illustrate the difference between our proposed method and MACN, we compared the two methods in four scenarios of Task1, as shown in Fig. \ref{fig_macn_ours} For scenario (a), both methods can direct the UGV to reach the target point, and both of them take 16 time steps, but it can be seen that their paths are different. For scenario (b), both methods can reach the target point, in which our method uses 16 time steps and MACN uses 25 time steps. Showing that in some scenarios, our methods can save time. Moreover, through analysis of motion command sequence, it can be found that the path planned by the MACN has collided with obstacles for some time, but our method hasn't collided with obstacles. For scenario (c), our method reaches the target point. In MACN, the UGV moves downward from the starting point. After encountering the middle obstacles, it doesn't change its direction of movement, and fails to reach the target point within the limited time step (we assume that if UGV doesn't reach the target point within 40 time steps, it is considered as a failure). For scenario (d), neither our method nor the MACN has reached the target point. The reason is that when the UGV moves downward and encounters the middle obstacles, the downward movement command is always used, resulting in the target point can't appear in the UAV's FOV. Therefore, we method can't jump to MCGN-II, so motion planning can't be performed. For scenario (e) with four obstacles, the MACN has the same path and time step as our method, and neither of them collided with the obstacles. Scenario (f) and (b) are similar. Both methods can reach the target point, but our method takes less time steps and doesn't collide with obstacles. Scenario (g) is similar to (b) and (f). Both methods can reach the target point. Among them, our method uses 17 time steps. The MACN issues an incorrect command when it is about to reach the target point, which leads to collision with obstacles, and MACN utilizes 18 time steps in total. For scenario (h), the MACN did not reach the target point. When it reached the position of the blue dot, the MACN instructed the UGV to move upward and collide with the obstacle. Although our method has reached the target point, it can be seen that our method doesn't get a good path, because there is a turn back path which is useless.
	
	Through the above analysis, one can conclude that our method effectively improves the success rate and reduce the situation of collision with obstacles compared with MACN. In most scenarios, our proposed method can direct UGV to reach the target point.
	
	\section{CONCLUSIONS}
	This paper has proposed a motion planning algorithm for the heterogeneous unmanned system under partial observation of environment from UAV. The algorithm consists of two parts. In the perception part, we have proposed GMGN, which can identify obstacles and passable parts in the pictures taken by UAVs. In the decision-making part, MCGN has been proposed. By the addition of memory mechanism, our methods can remember import landmarks in the exploration process, to obtain planning capabilities under partial observation from UAVs compared to similar state-of-the-art approaches.
	
	We evaluate our proposed algorithm by comparing with baseline algorithms. The results demonstrate that our method can effectively plan the motion of heterogeneous unmanned systems and achieve a relatively high success rate. Our future work will focus on the further improvement of accuracy rate and utilizing a UAV to assist multiple UGVs for motion planning.

	
	
	
	
	
	\bibliographystyle{IEEEtran}
	\bibliography{ref} 

\begin{thebibliography}{10}
\providecommand{\url}[1]{#1}
\csname url@rmstyle\endcsname
\providecommand{\newblock}{\relax}
\providecommand{\bibinfo}[2]{#2}
\providecommand\BIBentrySTDinterwordspacing{\spaceskip=0pt\relax}
\providecommand\BIBentryALTinterwordstretchfactor{4}
\providecommand\BIBentryALTinterwordspacing{\spaceskip=\fontdimen2\font plus
\BIBentryALTinterwordstretchfactor\fontdimen3\font minus
  \fontdimen4\font\relax}
\providecommand\BIBforeignlanguage[2]{{%
\expandafter\ifx\csname l@#1\endcsname\relax
\typeout{** WARNING: IEEEtran.bst: No hyphenation pattern has been}%
\typeout{** loaded for the language `#1'. Using the pattern for}%
\typeout{** the default language instead.}%
\else
\language=\csname l@#1\endcsname
\fi
#2}}

\bibitem{01}
E.~Mueggler, M.~Faessler, F.~Fontana, and D.~Scaramuzza, ``Aerial-guided
  navigation of a ground robot among movable obstacles,'' in \emph{2014 IEEE
  International Symposium on Safety, Security, and Rescue Robotics
  (2014)}.\hskip 1em plus 0.5em minus 0.4em\relax IEEE, 2014, pp. 1--8.

\bibitem{02}
A.~Lakas, B.~Belkhouche, O.~Benkraouda, A.~Shuaib, and H.~J. Alasmawi, ``A
  framework for a cooperative uav-ugv system for path discovery and planning,''
  in \emph{2018 International Conference on Innovations in Information
  Technology (IIT)}.\hskip 1em plus 0.5em minus 0.4em\relax IEEE, 2018, pp.
  42--46.

\bibitem{03}
J.~Li, G.~Deng, C.~Luo, Q.~Lin, Q.~Yan, and Z.~Ming, ``A hybrid path planning
  method in unmanned air/ground vehicle (uav/ugv) cooperative systems,''
  \emph{IEEE Transactions on Vehicular Technology}, vol.~65, no.~12, pp.
  9585--9596, 2016.

\bibitem{04}
J.~H. Kim, J.-W. Kwon, and J.~Seo, ``Multi-uav-based stereo vision system
  without gps for ground obstacle mapping to assist path planning of ugv,''
  \emph{Electronics Letters}, vol.~50, no.~20, pp. 1431--1432, 2014.

\bibitem{kashino2019aerial}
Z.~Kashino, G.~Nejat, and B.~Benhabib, ``Aerial wilderness search and rescue
  with ground support,'' \emph{Journal of Intelligent \& Robotic Systems}, pp.
  1--17, 2019.

\bibitem{papachristos2014power}
C.~Papachristos and A.~Tzes, ``The power-tethered uav-ugv team: A collaborative
  strategy for navigation in partially-mapped environments,'' in \emph{22nd
  Mediterranean Conference on Control and Automation}.\hskip 1em plus 0.5em
  minus 0.4em\relax IEEE, 2014, pp. 1153--1158.

\bibitem{peterson2018online}
J.~Peterson, H.~Chaudhry, K.~Abdelatty, J.~Bird, and K.~Kochersberger, ``Online
  aerial terrain mapping for ground robot navigation,'' \emph{Sensors},
  vol.~18, no.~2, p. 630, 2018.

\bibitem{13}
A.~Graves, G.~Wayne, and I.~Danihelka, ``Neural turing machines,'' \emph{arXiv
  preprint arXiv:1410.5401}, 2014.

\bibitem{14}
{\L}.~Kaiser and I.~Sutskever, ``Neural gpus learn algorithms,'' \emph{arXiv
  preprint arXiv:1511.08228}, 2015.

\bibitem{15}
W.~Zaremba and I.~Sutskever, ``Reinforcement learning neural turing
  machines-revised,'' \emph{arXiv preprint arXiv:1505.00521}, 2015.

\bibitem{16}
K.~Kurach, M.~Andrychowicz, and I.~Sutskever, ``Neural random-access
  machines,'' \emph{arXiv preprint arXiv:1511.06392}, 2015.

\bibitem{17}
E.~Grefenstette, K.~M. Hermann, M.~Suleyman, and P.~Blunsom, ``Learning to
  transduce with unbounded memory,'' in \emph{Advances in neural information
  processing systems}, 2015, pp. 1828--1836.

\bibitem{18}
A.~Joulin and T.~Mikolov, ``Inferring algorithmic patterns with stack-augmented
  recurrent nets,'' in \emph{Advances in neural information processing
  systems}, 2015, pp. 190--198.

\bibitem{19}
A.~Graves, G.~Wayne, M.~Reynolds, T.~Harley, I.~Danihelka,
  A.~Grabska-Barwi{\'n}ska, S.~G. Colmenarejo, E.~Grefenstette, T.~Ramalho,
  J.~Agapiou, \emph{et~al.}, ``Hybrid computing using a neural network with
  dynamic external memory,'' \emph{Nature}, vol. 538, no. 7626, pp. 471--476,
  2016.

\bibitem{12}
A.~Paszke, A.~Chaurasia, S.~Kim, and E.~Culurciello, ``Enet: A deep neural
  network architecture for real-time semantic segmentation,'' \emph{arXiv
  preprint arXiv:1606.02147}, 2016.

\bibitem{21}
A.~Tamar, Y.~Wu, G.~Thomas, S.~Levine, and P.~Abbeel, ``Value iteration
  networks,'' in \emph{Advances in Neural Information Processing Systems},
  2016, pp. 2154--2162.

\bibitem{22}
A.~Khan, C.~Zhang, N.~Atanasov, K.~Karydis, V.~Kumar, and D.~D. Lee, ``Memory
  augmented control networks,'' \emph{arXiv preprint arXiv:1709.05706}, 2017.

\bibitem{23}
S.~Li, Y.~Wan, P.~He, C.~Wang, J.~Sun, Y.~Zhang, X.~Li, and G.~Xiel, ``Heros: A
  simulation platform for heterogeneous robotic swarms,'' in \emph{2019 Chinese
  Control Conference (CCC)}.\hskip 1em plus 0.5em minus 0.4em\relax IEEE, 2019,
  pp. 7223--7228.

\end{thebibliography}
	
\end{document}